\documentclass[sigconf,nonacm]{acmart}
\usepackage{subcaption}
\usepackage{makecell}
\usepackage{multirow}

\graphicspath{{figures/}}

\AtBeginDocument{%
  }

\setcopyright{none}
\providecommand{\correspondingauthor}{}

\begin{document}

\title{Recovering Lost Details: Multi-Scale Frequency Compensation for Long-Term Time Series Forecasting}

\author{Runmin Zou}
\affiliation{%
  \institution{Central South University}
  \city{Changsha}
  \state{Hunan}
  \country{China}
}
\email{rmzou@csu.edu.cn}

\author{Siyi Xie}
\affiliation{%
  \institution{Central South University}
  \city{Changsha}
  \state{Hunan}
  \country{China}}
\email{244611064@csu.edu.cn}

\author{Yaohui Huang}
\affiliation{%
  \institution{Central South University}
  \city{Changsha}
  \state{Hunan}
  \country{China}
}
\email{yaohuihuang@csu.edu.cn}

\author{Yun Wang}
\correspondingauthor
\affiliation{%
  \institution{Central South University}
  \city{Changsha}
  \state{Hunan}
  \country{China}}
\email{wangyun19@csu.edu.cn}


\begin{abstract}
  Long-term time series forecasting has made significant progress by leveraging multi-scale information to capture hierarchical temporal patterns and model long-range dependencies. However, temporal downsampling in existing multi-scale methods inevitably smooths detailed temporal fluctuations, and this information loss is further aggravated by their emphasis on dominant trends across scales, resulting in insufficiently expressive representations. To address this, we propose a Multi-Scale Wavelet Mixing (MWMixer) model, which incorporates a Bidirectional Frequency-Bands Mixing strategy to recover lost temporal details across scales, enabling complementary cross-scale information interactions. Then, a Dynamic Scale-Adaptive Fusion module learns time-varying weights for each scale to fuse multi-scale forecasts into the final prediction, enhancing the flexibility of multi-scale aggregation. In addition, a cross-scale consistency loss aligns each coarse-scale prediction with the interval-averaged fine-scale outputs, while a multi-scale supervision loss enforces prediction accuracy at each scale, promoting consistent learning across scales. Extensive experiments on seven real-world datasets demonstrate that MWMixer achieves competitive performance in long-term forecasting.
\end{abstract}




\maketitle

\section{Introduction}

Time series forecasting has become a critical research problem due to its broad applications in finance~\cite{finance}, energy~\cite{energy,11563890}, weather forecasting~\cite{liu2025timecmaweather,miao2024lessweather}, and hydrology~\cite{shuiwen}.

\begin{figure}[t]
  \centering
  \includegraphics[width=\columnwidth]{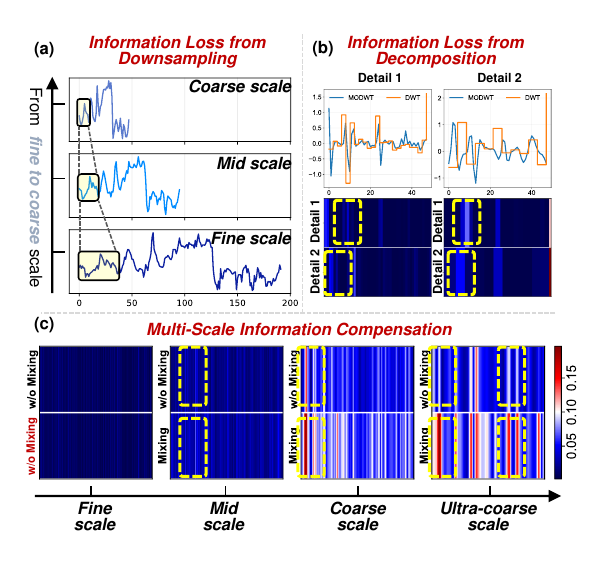}
  \caption{Motivation of the proposed framework. (a) Downsampling preserves global trends from fine to coarse scale but erodes fine-grained local details. (b) Unlike Discrete Wavelet Transform (DWT) sequence reduction, the Maximal Overlap DWT (MODWT) maintains full temporal resolution. (c) Energy maps reveal that frequency-domain compensation Mixing successfully restores information lost during downsampling compared to the pre-compensation state w/o Mixing.}
  \Description{A three-part motivation diagram. Panel (a) depicts a fine-scale time series becoming progressively smoother after downsampling to coarser scales, illustrating the loss of local fluctuations. Panel (b) contrasts DWT, whose coefficients become shorter after decomposition, with MODWT, whose coefficient sequences retain the original temporal resolution. Panel (c) compares coefficient-energy heatmaps before and after frequency-band mixing and shows the recovery of energy patterns weakened by downsampling.}
  \label{fig:motivation}
\end{figure}

Deep learning architectures, ranging from RNNs to Transformers like SegRNN~\cite{lin2023segrnn} and PatchTST~\cite{Yuqietal-2023-PatchTST} have fundamentally advanced time series forecasting. Yet, these methods predominantly operate on a single temporal scale, making it difficult to disentangle complex variations where evolutionary patterns manifest at distinct granularities. Multi-scale frameworks address this by utilizing downsampling to model global trends, but this approach suffers from \textbf{\textit{information loss from downsampling}}. As shown in Figure~\ref{fig:motivation}(a), the resolution reduction inevitably smooths out critical detail information, such as sudden fluctuations. While time-domain decomposition often struggles to model such non-stationary dynamics, frequency-domain analysis reveals underlying structures. Compared to the global view of the Fourier Transform, wavelet transforms offer a more comprehensive characterization of local fluctuations and detailed structures. However, applying the standard DWT introduces the \textbf{\textit{information loss from decomposition}}. As illustrated in Figure~\ref{fig:motivation}(b), the intrinsic downsampling operation of DWT leads to sequence shortening, resulting in coefficients that are too sparse for effective feature interaction on coarse scales. Ultimately, the compounding effect of these two sources of information loss leads to insufficiently expressive representations, severely limiting the model's ability to capture the intricate interplay between local fluctuations and global trends. 

To address the information loss arising from both downsampling and decomposition, we introduce the Multi-Scale Wavelet Mixing (MWMixer) framework. First, to overcome the information loss from decomposition, we employ MODWT, which maintains full sequence length and shift-invariance, enabling effective feature interaction at all scales. Second, to mitigate the information loss from downsampling, we propose a Bidirectional Frequency-Bands Mixing (BFBM) module. As evidenced in Figure~\ref{fig:motivation}(c), this module explicitly compensates for the lost details through precise frequency-domain mixing by injecting fine-scale information into coarse-scale representations while preserving global trends. Additionally, our Dynamic Scale-Adaptive Fusion (DSAF) module learns time-varying weights to adaptively fuse forecasts, replacing rigid static aggregation. Finally, a Scale Loss strategy combining cross-scale consistency and multi-scale supervision ensures structural coherence across the scales.
In summary, our main contributions are as follows:
\begin{itemize}
\item We propose an MWMixer framework for long-term time series forecasting, which effectively captures complex temporal patterns from the integration of time-domain downsampling and frequency-domain decomposition in a unified architecture.
\item We employ MODWT to prevent information loss from decomposition. Crucially, we design the BFBM module to explicitly compensate for information loss from downsampling by recovering lost details via bidirectional interaction. Then, DSAF and Scale Loss are introduced to ensure adaptive fusion and structural consistency.
\item We conduct extensive experiments on seven mainstream time-series benchmarks. Results demonstrate that MWMixer achieves highly competitive forecasting accuracy.
\end{itemize}

\section{Related Work}

\subsection{Time Series Forecasting}

Deep learning has advanced time series forecasting. Early research focused on RNNs~\cite{deepar,lin2023segrnn} and CNNs~\cite{liu2022SCINet,wu2023timesnet}, while subsequent attention-based architectures, such as Informer~\cite{haoyietal-informer-2021}, Autoformer~\cite{wu2021autoformer}, and PatchTST~\cite{Yuqietal-2023-PatchTST}, became dominant for their ability to capture long-range dependencies. Despite this success, the high computational cost of Transformers has sparked a resurgence of lightweight MLP-based models. Notably, DLinear~\cite{Zeng2022dlinear}, TiDE~\cite{das2023tide}, and TSMixer~\cite{chen2023tsmixer} demonstrate that simple linear mappings or channel-mixing structures can achieve state-of-the-art performance with superior efficiency. Additionally, frequency-domain methods like FEDformer~\cite{zhou2022fedformer} and FourierGNN~\cite{yi2023fouriergnn} leverage spectral projections to model global properties. However, a critical common limitation persists across these paradigms: they predominantly operate under a single-scale perspective. By processing data solely at the original resolution, these methods often struggle to disentangle complex hierarchical variations where evolutionary patterns manifest at distinct temporal granularities.

\subsection{Multi-Scale Modeling}

To overcome the inherent limitations of single-scale modeling, multi-scale architectures utilize hierarchical representations to capture both long-term trends and short-term fluctuations simultaneously. Pyraformer~\cite{liu2022pyraformer} constructs a pyramidal attention mechanism to build a multi-resolution hierarchy, efficiently aggregating historical information with linear complexity. Similarly, SCINet~\cite{liu2022SCINet} adopts a recursive downsampling-interaction framework, utilizing a binary tree structure to extract distinct temporal features across varying resolutions and exponentially expand the receptive field. Scaleformer~\cite{shabani2023scaleformer}  further refines this by employing an iterative framework that enforces consistency from coarse to fine scales.

Recent advancements have shifted towards explicit pattern disentanglement. MICN~\cite{micn}  employs multi-scale isometric convolutions to separately model local patterns and global correlations. Furthermore, TimeMixer~\cite{wang2023timemixer} and AMD~\cite{hu2025amd} construct multi-scale representations through
average pooling and mix temporal patterns across the resulting scales.
Although effective for trend modeling, pooling acts as a low-pass filter
and can attenuate high-frequency details. MWMixer also uses averaging to
construct the scale hierarchy, but further applies intra-scale MODWT to
decouple detail and approximation coefficients. BFBM then propagates
detail coefficients from fine to coarse scales and approximation
coefficients in the reverse direction, compensating attenuated details
while enriching low-frequency representations. Thus, unlike their
time-domain multi-scale modeling, MWMixer explicitly performs
frequency-band compensation before adaptive forecast fusion.

\begin{figure*}[t]
  \centering
  \includegraphics[width=\textwidth]{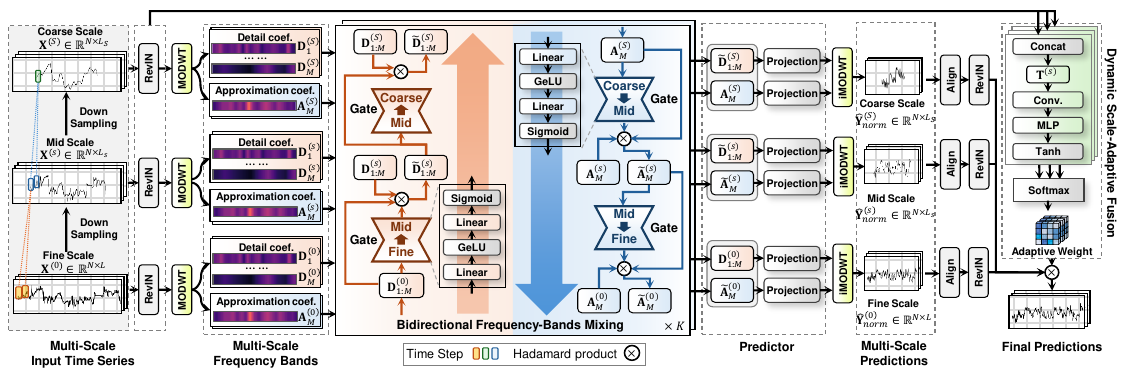}
  \caption{Overview of the proposed model architecture. The model processes multi-scale time-series inputs and decomposes them in the frequency domain via MODWT, enabling explicit multi-band representations. The BFBM module captures cross-scale interactions from coarse to fine and vice versa, followed by DSAF to integrate multi-scale predictions.}
  \Description{A left-to-right architecture diagram of MWMixer. The input sequence is downsampled into multiple temporal scales, and each scale is decomposed by MODWT into approximation and detail coefficients. Bidirectional Frequency-Bands Mixing exchanges information between adjacent scales in both coarse-to-fine and fine-to-coarse directions. The reconstructed scale-specific representations are processed by predictors, and Dynamic Scale-Adaptive Fusion combines their forecasts using time-varying scale weights to produce the final prediction.}
  \label{fig:model_architecture}
\end{figure*}

\section{Method}

 Given a multivariate time series input $\mathbf{X}=(\mathbf{x}_{t-L+1}, \dots, \mathbf{x}_{t}) \in \mathbb{R}^{N\times L} $ with a look-back window $L$, the goal is to predict the subsequent future sequence $\mathbf{Y}=(\mathbf{x}_{t+1}, \dots, \mathbf{x}_{t+H}) \in \mathbb{R}^{N\times H}$ with a horizon window $H$, where $\mathbf{x}_{i} \in \mathbb{R}^{N}$ represents the observation vector of $N$ variables at time step $i$, serving as the fundamental unit for both historical observation and future prediction.

\subsection{Overall Architecture}
An overview of the proposed framework is shown in Figure~\ref{fig:model_architecture}. To explicitly capture temporal dependencies at varying granularities, we first construct a multi-scale input representation. Formally, given a multivariate input series $\mathbf{X}$, we generate a set of multi-scale inputs $\{\mathbf{X}^{(0)}, \mathbf{X}^{(1)}, \ldots, \mathbf{X}^{(S)}\}$ via average-pooling downsampling. Here, $\mathbf{X}^{(0)} = \mathbf{X}$ denotes the original input. The input $\mathbf{X}^{(s)}$ at scale $s$ is derived by applying $\mathrm{AvgPooling}(\cdot)$ with a non-overlapping window size $w$ to the representation at the previous scale $\mathbf{X}^{(s-1)}$:
\begin{equation}
\resizebox{.91\linewidth}{!}{$
\displaystyle
\mathbf{X}^{(s)} = \mathrm{AvgPooling}\!\left(\mathbf{X}^{(s-1)}\right) \in \mathbb{R}^{N \times L_s},
\quad s \in \{1,\ldots,S\},
$}
\label{eq:avgpool}
\end{equation}
where $L_s = \left\lfloor L / w^s \right\rfloor$ denotes the sequence length at scale $s$. This downsampling-based multi-scale mechanism effectively expands the receptive field, extracting dominant long-term trends while filtering out high-frequency noise.

To mitigate distribution shift and non-stationarity, we adopt Reversible Instance Normalization (RevIN)~\cite{kim2021reversible}, applied independently to each instance at every scale $s$. This independent application is crucial because each scale represents distinct temporal patterns and exhibits unique data distributions. Specifically, the input $\mathbf{X}^{(s)}$ is normalized as $\tilde{\mathbf{X}}^{(s)}$ before entering the network. This instance-wise normalization will be inverted during the prediction stage at each scale to restore the data to their original magnitude.

Next, to disentangle complex temporal patterns at different granularities, we apply MODWT independently at each scale to map the normalized series into the frequency domain. We adopt MODWT instead of the standard DWT because it is inherently shift-invariant and preserves signal length, thereby reducing information loss during decomposition. For each scale s, we perform an $M$-level MODWT decomposition to obtain a set of detail coefficients $\mathbf{D}$ and approximation coefficients $\mathbf{A}$:
\begin{equation}
\{\mathbf{D}_1^{(s)},\mathbf{D}_2^{(s)},\ldots,\mathbf{D}_M^{(s)},\mathbf{A}_M^{(s)}\}={\mathrm{MODWT}}(\tilde{\mathbf{X}}^{(s)}),
\label{eq:MODWT}
\end{equation}
where $\mathbf{D}_m^{(s)}\in \mathbb{R}^{N\times L_s}$ denotes the high-frequency detail coefficients at level $m$ for scale $s$, while $\mathbf{A}_M^{(s)}\in \mathbb{R}^{N{\times L}_s}$ denotes the low-frequency approximation coefficients at level $M$. This design applies time-domain downsampling followed by frequency-domain decomposition to organize temporal dynamics into a unified tensor defined over scale, frequency, and time. The downsampling operation enlarges the receptive field to capture long-range dependencies, while simultaneously acting as a low-pass filter that improves the signal-to-noise ratio. 

Then, we embed the decomposed coefficients into a latent space with dimension $d$ and process them using $K$ stacked layers of the BFBM module to capture cross-scale correlations. Finally, they are projected back to wavelet coefficients via a multi-layer perceptron (MLP) and transformed to the multi-scale time domain through the inverse MODWT (iMODWT), which can be written as:
\begin{equation}
{\hat{\mathbf{Y}}}_{norm}^{(s)}={\mathrm{iMODWT}}\{\hat{\mathbf{D}}_1^{(s)},\hat{\mathbf{D}}_2^{(s)},\ldots,\hat{\mathbf{D}}_M^{(s)},\hat{\mathbf{A}}_M^{(s)}\},
\label{eq:iMODWT}
\end{equation}
where $\hat{\mathbf{D}}_M^{(s)}$ and $\hat{\mathbf{A}}_M^{(s)}$ denote the predicted level-$M$ detail coefficients and approximation coefficients at scale $s$, respectively. Subsequently, we upsample the normalized prediction ${\hat{\mathbf{Y}}}_{norm}^{(s)}$ to align with the finest-scale length and apply reversible instance denormalization to obtain the scale-wise prediction ${\hat{\mathbf{Y}}}^{(s)}$ at the original magnitude. 

Finally, we aggregate these predictions via the DSAF module to produce the final output.

\subsection{Bidirectional Frequency-Bands Mixing}

Although the nested frequency-domain decomposition provides a disentangled representation of temporal features, processing each scale in isolation ignores the inherent inter-scale correlations that are crucial for holistic modeling. Theoretically, as established in Theorem \ref{thm:modwt_separation}, MODWT coefficients at different levels occupy spectrally distinct frequency intervals, effectively creating complementary views of the underlying dynamics. While this spectral separation is beneficial for analysis, holistic modeling requires bridging these distinct views to capture the full structural dependencies. To achieve this, we introduce the BFBM module to enable selective information flow across scales.
\begin{theorem}
{(Spectral Separation and Energy Conservation of MODWT).}
Given a time series $\mathbf{X}$, the MODWT decomposition yields coefficients $\{\mathbf{D}_1, \dots, \mathbf{D}_M, \mathbf{A}_M\}$ that are spectrally disjoint. The detail coefficients $\mathbf{D}_m$ and approximation coefficients $\mathbf{A}_M$ occupy mutually exclusive frequency passbands, forming a complete partition of the signal's energy such that $\|\mathbf{X}\|_F^2 = \sum_{m=1}^{M}\|\mathbf{D}_m\|_F^2 + \|\mathbf{A}_M\|_F^2$. Proof is provided in Appendix A.
\label{thm:modwt_separation}
\end{theorem}

However, wavelet coefficients at different decomposition levels often exhibit significant energy discrepancies. Consequently, naive summation is numerically unstable and prone to feature dominance, where high-energy low-frequency components suppress fine-grained details. 

We adopt a context-aware gating mechanism to mitigate this as illustrated in Figure~\ref{fig:model_architecture}. Specifically, an MLP dynamically generates a learnable weight tensor that acts as a soft normalization factor. The MLP for an arbitrary input matrix $\mathbf{Z}$ is formulated as:
\begin{equation}
    \mathrm{MLP}(\mathbf{Z}) = \mathbf{W}_2^{(n)}(\text{GELU}(\mathbf{W}_1^{(n)}\mathbf{Z} + \mathbf{b}_1)) + \mathbf{b}_2,
    \label{eq:mlp_def}
\end{equation}
where $\mathbf{W}_1^{(n)}, \mathbf{W}_2^{(n)}$ are learnable weights and $\mathbf{b}_1, \mathbf{b}_2$ are bias terms. It rescales source coefficients to compatible magnitudes, ensuring numerical stability while selectively suppressing irrelevant fluctuations during the mixing process. Built on the above mechanism, the module performs bidirectional interactions along two complementary information-flow paths. 

\paragraph{Fine-to-Coarse Detail Path.} The fine-to-coarse detail propagation path focuses on the detail coefficients $\mathbf{D}_m^{(s)}$ , which typically encode rapid fluctuations and local transients. However, at coarser scales, 
the low-pass effect of downsampling can substantially attenuate or even overwhelm high-frequency details. To 
prevent critical local events from being lost during scale downsampling, we selectively inject fine-scale (scale 
$s$) detail information into the coarse-scale (scale $s+1$) representation. Specifically, for the level-$m$ detail coefficients, we compute a context-aware gating vector based on the processed details from the finer scale:
\begin{equation}
\mathbf{G}_{Dm}^{(s)} = \sigma\left(\mathrm{MLP}_D\left(\tilde{\mathbf{D}}_m^{(s)}\right)\right),
\label{eq:weightd}
\end{equation}
where $\sigma$ is the sigmoid function. Then, update the coarse-scale detail coefficients via:
\begin{equation}
\tilde{\mathbf{D}}_m^{(s+1)} \leftarrow \left(1 - \mathbf{G}_{Dm}^{(s)}\right) \odot \mathbf{D}_m^{(s+1)} + \mathbf{G}_{Dm}^{(s)} \odot \tilde{\mathbf{D}}_m^{(s)},
\label{eq:gated}
\end{equation}
where $\odot$ denotes element-wise multiplication and $\tilde{\mathbf{D}}_m^{(s)}$ represents the updated detail representation at scale $s$. Note that for the initial scale ($s=0$), we define $\tilde{\mathbf{D}}_m^{(0)} = \mathbf{D}_m^{(0)}$. 

\paragraph{Coarse-to-Fine Approximation Path.} Conversely, the coarse-to-fine path focuses on approximation coefficients $\mathbf{A}_m^{(s)}$, which primarily encode long-term trends and global structure. Since coarser scales offer larger receptive fields, they often provide more robust estimates of the overall trend. We therefore propagate the trend information from the coarse scale (scale $s+1$) back to the fine scale (scale $s$), serving as a global prior for fine-scale representation learning. Accordingly, the gating weights and update rules are defined as:
\begin{equation}
\mathbf{G}_{Am}^{(s)} = \sigma\left(\mathrm{MLP}_A\left(\tilde{\mathbf{A}}_m^{(s+1)}\right)\right).
\label{eq:weighta}
\end{equation}

Then, update the fine-scale approximation coefficients via:
\begin{equation}
\tilde{\mathbf{A}}_m^{(s)} \leftarrow \left(1 - \mathbf{G}_{Am}^{(s)}\right) \odot \mathbf{A}_m^{(s)} + \mathbf{G}_{Am}^{(s)} \odot \tilde{\mathbf{A}}_m^{(s+1)},
\label{eq:gatea}
\end{equation}
where $\tilde{\mathbf{A}}_m^{(s)}$ represents the updated approximation representation at scale $s$. Note that for the coarsest scale ($s=S$), we define $\tilde{\mathbf{A}}_m^{(S)} = \mathbf{A}_m^{(S)}$. This trend allows local representations to remain consistent with the global trend while preserving their ability to capture details.

In summary, this bidirectional interaction enables the framework to simultaneously preserve global trends while recovering lost details across varying scales.

\subsection{Dynamic Scale-Adaptive Fusion}

While the multi-scale architecture generates diverse candidate forecasts, effectively aggregating them requires 
reconciling inherent trade-offs across granularities. Fine-grained scales excel at capturing short-term, 
high-frequency fluctuations but are susceptible to error accumulation over longer horizons. In contrast, 
coarse-grained scales offer robust modeling of long-term trends but often lack the resolution to depict 
fine-grained local variations. 

However, conventional methods typically rely on static averaging, failing to account for the time-varying reliability of these scales. To address this 
limitation, we introduce the DSAF module, as illustrated in Figure~\ref{fig:model_architecture}.
Functioning as a meta-learner, DSAF dynamically evaluates the 
contribution of each scale at every prediction step, synthesizing the final forecast by adaptively 
weighting scales based on their estimated reliability.

Formally, after iMODWT, we obtain a set of multi-scale predicted sequences $\{\hat{\mathbf{Y}}_{norm}^{(0)}, \hat{\mathbf{Y}}_{norm}^{(1)}, \ldots, \hat{\mathbf{Y}}_{norm}^{(S)}\}$, where $\hat{\mathbf{Y}}_{norm}^{(s)} \in \mathbb{R}^{N \times H_s}$ denotes the normalized prediction at scale $s$ with horizon length $H_s = \left\lfloor H / w^s \right\rfloor$. For each scale $s$, prediction reliability is not uniformly distributed over the forecasting horizon. 

To quantify this, we first construct a complete trajectory $\mathbf{T}^{(s)}$ by concatenating the downsampled normalized historical context $\tilde{\mathbf{X}}^{(s)}$ with the raw prediction $\hat{\mathbf{Y}}_{norm}^{(s)}$:
\begin{equation}
\mathbf{T}^{(s)} = \mathrm{Concat}\left(\tilde{\mathbf{X}}^{(s)}, \hat{\mathbf{Y}}_{norm}^{(s)}\right).
\label{eq:concat}
\end{equation}

Subsequently, to capture the temporal continuity and transition patterns between the historical context and the forecast, we employ a lightweight 1D convolutional network to extract features from the trajectory $\mathbf{T}^{(s)}$. These features are mapped via an MLP to generate a confidence score $e_{k,s}$ for each future time step $k$ at scale $s$.
Then, to determine the specific contribution of each scale, we apply a softmax normalization to the confidence scores along the scale dimension:
\begin{equation}
w_{k}^{(s)} = \frac{\exp(e_{k,s})}{\sum_{j=0}^{S} \exp(e_{k,j})},
\label{eq:softmax}
\end{equation}
where $w_{k}^{(s)} \in [0,1]$ denotes the normalized importance weight of scale $s$ at time step $k$. $\mathbf{W}^{(s)} = [w_{1}^{(s)}, \dots, w_{H}^{(s)}]^\top$ is the time-varying weight vector for scale $s$. 

Consequently, the final prediction $\hat{\mathbf{Y}}^{final}$ is obtained by the adaptive weighted aggregation of the upsampled and denormalized forecasts:
\begin{equation}
\hat{\mathbf{Y}}^{final} = \sum\nolimits_{s=0}^{S} \mathbf{W}^{(s)} \odot \hat{\mathbf{Y}}^{(s)}.
\label{eq:aggregate1}
\end{equation}

This mechanism adaptively selects the most reliable scale for each specific time step across the forecasting horizon.

\begin{table*}[t]
\centering
\caption{Multivariate long-term forecasting results (MSE/MAE) under prediction horizons $H \in \{96, 192, 336, 720\}$ with a fixed look-back length $L=96$. ``Avg'' denotes the average performance across all horizons. Lower values indicate better performance. The best results are highlighted in \textbf{bold}, and the second-best results are \underline{underlined}. The ``1st count'' row reports the number of best results achieved by each method in this table.}
\label{tab:all_results}

\resizebox{\textwidth}{!}{
\begin{tabular}{c|c|*{9}{cc|}cc}
\toprule
\multicolumn{2}{c|}{Method}
& \multicolumn{2}{c|}{\makecell[c]{MWMixer\\(ours)}}
& \multicolumn{2}{c|}{\makecell[c]{CASA\\(2025)}}
& \multicolumn{2}{c|}{\makecell[c]{DUET\\(2025)}}
& \multicolumn{2}{c|}{\makecell[c]{AMD\\(2025)}}
& \multicolumn{2}{c|}{\makecell[c]{TimeMixer\\(2024)}}
& \multicolumn{2}{c|}{\makecell[c]{SparseTSF\\(2024)}}
& \multicolumn{2}{c|}{\makecell[c]{iTransformer\\(2024)}}
& \multicolumn{2}{c|}{\makecell[c]{FilterNet\\(2024)}}
& \multicolumn{2}{c|}{\makecell[c]{FITS\\(2024)}}
& \multicolumn{2}{c}{\makecell[c]{MICN\\(2023)}}
\\
\cline{1-22}
\multicolumn{2}{c|}{Metric}
& MSE & MAE & MSE & MAE & MSE & MAE & MSE & MAE & MSE & MAE
& MSE & MAE & MSE & MAE & MSE & MAE & MSE & MAE & MSE & MAE
\\
\midrule

\multirow{5}{*}{\rotatebox[origin=c]{90}{ETTh1}}
& 96  &  \textbf{0.368} & \textbf{0.389} & 0.382 & 0.404 & 0.400 & 0.405 & 0.383 & 0.393 & \underline{0.372} & 0.398 & 0.385 & \underline{0.390} & 0.385 & 0.404 & 0.382 & 0.402 & 0.384 & 0.392 & 0.421 & 0.431 \\
& 192 &  \underline{0.416} & \textbf{0.417} & 0.427 & 0.428 & 0.430 & 0.425 & 0.435 & 0.422 & 0.437 & 0.427 & \textbf{0.414} & \underline{0.420} & 0.439 & 0.434 & 0.430 & 0.429 & 0.434 & 0.421 & 0.474 & 0.487 \\
& 336 &  \textbf{0.450} & \textbf{0.436} & 0.474 & 0.451 & 0.479 & 0.452 & 0.475 & 0.443 & 0.489 & 0.457 & 0.476 & \underline{0.439} & 0.492 & 0.463 & \underline{0.472} & 0.451 & 0.475 & 0.442 & 0.569 & 0.551 \\
& 720 &  \textbf{0.440} & \textbf{0.449} & 0.484 & 0.475 & 0.527 & 0.494 & 0.479 & 0.464 & 0.479 & 0.471 & \underline{0.461} & \underline{0.454} & 0.509 & 0.493 & 0.481 & 0.473 & 0.462 & 0.459 & 0.770 & 0.672 \\
& Avg &  \textbf{0.418} & \textbf{0.422} & 0.441 & 0.439 & 0.459 & 0.444 & 0.443 & 0.430 & 0.444 & 0.438 & \underline{0.434} & \underline{0.425} & 0.456 & 0.448 & 0.441 & 0.438 & 0.438 & 0.428 & 0.558 & 0.535 \\
\midrule

\multirow{5}{*}{\rotatebox[origin=c]{90}{ETTh2}}
& 96  &  \textbf{0.282} & \textbf{0.335} & 0.296 & 0.347 & 0.295 & 0.345 & \underline{0.285} & \underline{0.337} & 0.292 & 0.343 & 0.302 & 0.346 & 0.300 & 0.349 & 0.293 & 0.343 & 0.289 & 0.338 & 0.299 & 0.364 \\
& 192 &  \textbf{0.362} & \textbf{0.386} & 0.375 & 0.397 & \underline{0.367} & \underline{0.388} & 0.387 & 0.394 & 0.373 & 0.394 & 0.384 & 0.395 & 0.380 & 0.397 & 0.374 & 0.396 & 0.375 & 0.390 & 0.441 & 0.454 \\
& 336 &  \underline{0.405} & \underline{0.422} & 0.420 & 0.433 & 0.411 & \underline{0.422} & 0.416 & 0.427 & \textbf{0.387} & \textbf{0.415} & 0.421 & 0.427 & 0.425 & 0.433 & 0.417 & 0.430 & 0.413 & 0.424 & 0.654 & 0.567 \\
& 720 &  \textbf{0.414} & \textbf{0.435} & 0.430 & 0.446 & \underline{0.416} & \underline{0.436} & 0.422 & 0.441 & 0.417 & 0.438 & 0.420 & 0.437 & 0.429 & 0.445 & 0.449 & 0.460 & \underline{0.416} & \textbf{0.435} & 0.956 & 0.716 \\
& Avg &  \textbf{0.366} & \textbf{0.395} & 0.380 & 0.405 & 0.372 & 0.397 & 0.377 & 0.399 & \underline{0.367} & 0.397 & 0.381 & 0.401 & 0.383 & 0.406 & 0.383 & 0.407 & 0.373 & \underline{0.396} & 0.587 & 0.525 \\
\midrule

\multirow{5}{*}{\rotatebox[origin=c]{90}{ETTm1}}
& 96  &  \textbf{0.310} & \textbf{0.351} & 0.319 & 0.358 & 0.326 & \underline{0.356} & 0.322 & 0.359 & 0.320 & 0.358 & 0.356 & 0.375 & 0.343 & 0.377 & 0.321 & 0.361 & 0.353 & 0.374 & \underline{0.316} & 0.362 \\
& 192 &  \textbf{0.353} & \textbf{0.376} & 0.376 & 0.390 & 0.368 & \underline{0.379} & 0.367 & 0.381 & 0.365 & 0.382 & 0.394 & 0.392 & 0.381 & 0.395 & 0.367 & 0.387 & 0.391 & 0.392 & \underline{0.363} & 0.390 \\
& 336 &  \textbf{0.387} & \textbf{0.400} & 0.402 & 0.407 & 0.401 & \underline{0.401} & 0.399 & \underline{0.401} & \underline{0.394} & 0.407 & 0.425 & 0.413 & 0.417 & 0.417 & 0.401 & 0.409 & 0.424 & 0.414 & 0.408 & 0.426 \\
& 720 &  \textbf{0.443} & \textbf{0.435} & \underline{0.459} & 0.444 & 0.463 & \underline{0.437} & 0.465 & \underline{0.437} & 0.460 & 0.448 & 0.487 & 0.448 & 0.488 & 0.457 & 0.477 & 0.448 & 0.480 & 0.448 & 0.481 & 0.476 \\
& Avg &  \textbf{0.373} & \textbf{0.390} & 0.389 & 0.399 & 0.389 & \underline{0.393} & 0.388 & 0.394 & \underline{0.384} & 0.398 & 0.415 & 0.407 & 0.407 & 0.411 & 0.391 & 0.401 & 0.412 & 0.407 & 0.392 & 0.413 \\
\midrule

\multirow{5}{*}{\rotatebox[origin=c]{90}{ETTm2}}
& 96  &  \textbf{0.173} & \textbf{0.254} & 0.176 & 0.258 & \textbf{0.173} & \textbf{0.254} & 0.180 & 0.264 & \underline{0.175} & \underline{0.257} & 0.184 & 0.267 & 0.185 & 0.272 & \underline{0.175} & 0.258 & 0.182 & 0.265 & 0.179 & 0.275 \\
& 192 &  \textbf{0.236} & \textbf{0.296} & 0.243 & 0.302 & \underline{0.238} & \underline{0.298} & 0.243 & 0.304 & 0.239 & 0.300 & 0.248 & 0.305 & 0.253 & 0.313 & 0.240 & 0.301 & 0.246 & 0.304 & 0.307 & 0.376 \\
& 336 &  \textbf{0.293} & \textbf{0.333} & \underline{0.296} & \underline{0.336} & 0.301 & 0.339 & 0.301 & 0.339 & 0.303 & 0.341 & 0.307 & 0.342 & 0.315 & 0.350 & 0.311 & 0.347 & 0.306 & 0.341 & 0.325 & 0.388 \\
& 720 &  \textbf{0.390} & \textbf{0.391} & 0.399 & 0.396 & 0.397 & \underline{0.395} & 0.402 & 0.396 & \underline{0.393} & \underline{0.395} & 0.407 & 0.398 & 0.413 & 0.406 & 0.414 & 0.405 & 0.407 & 0.397 & 0.502 & 0.490 \\
& Avg &  \textbf{0.273} & \textbf{0.319} & 0.279 & 0.323 & \underline{0.277} & \underline{0.322} & 0.282 & 0.326 & 0.278 & 0.323 & 0.287 & 0.328 & 0.292 & 0.335 & 0.285 & 0.327 & 0.285 & 0.326 & 0.328 & 0.382 \\
\midrule

\multirow{5}{*}{\rotatebox[origin=c]{90}{Weather}}
& 96  &  \underline{0.162} & \underline{0.206} & \textbf{0.158} & \underline{0.203} & \underline{0.161} & \textbf{0.199} & 0.181 & 0.226 & \underline{0.162} & 0.209 & 0.197 & 0.236 & 0.177 & 0.216 & \underline{0.162} & 0.207 & 0.165 & 0.211 & 0.167 & 0.235 \\
& 192 &  \textbf{0.208} & \textbf{0.249} & \underline{0.210} & 0.252 & 0.221 & 0.255 & 0.227 & 0.264 & \textbf{0.208} & 0.253 & 0.243 & 0.273 & 0.225 & 0.258 & \underline{0.210} & \underline{0.250} & 0.212 & 0.253 & 0.221 & 0.283 \\
& 336 &  \underline{0.264} & \textbf{0.289} & 0.266 & 0.293 & 0.274 & 0.292 & 0.281 & 0.301 & \textbf{0.263} & 0.292 & 0.292 & 0.308 & 0.282 & 0.299 & 0.265 & \underline{0.290} & 0.265 & 0.291 & 0.278 & 0.338 \\
& 720 &  0.343 & \textbf{0.340} & 0.353 & 0.346 & 0.347 & \textbf{0.340} & 0.355 & 0.348 & \underline{0.342} & 0.346 & 0.368 & 0.357 & 0.359 & 0.350 & \underline{0.342} & \textbf{0.340} & 0.345 & \underline{0.342} & \textbf{0.338} & 0.381 \\
& Avg &  \underline{0.244} & \textbf{0.271} & 0.246 & \underline{0.273} & 0.250 & \textbf{0.271} & 0.261 & 0.284 & \textbf{0.243} & 0.275 & 0.275 & 0.293 & 0.260 & 0.280 & \underline{0.244} & \textbf{0.271} & 0.246 & 0.274 & 0.251 & 0.309 \\
\midrule

\multirow{5}{*}{\rotatebox[origin=c]{90}{Exchange}}
& 96  &  \textbf{0.081} & \textbf{0.198} & 0.087 & 0.207 & 0.086 & 0.204 & 0.084 & 0.202 & \underline{0.083} & \underline{0.201} & 0.084 & \underline{0.201} & 0.086 & 0.206 & 0.091 & 0.211 & 0.088 & 0.207 & 0.102 & 0.235 \\
& 192 &  \textbf{0.165} & \textbf{0.289} & 0.177 & 0.300 & 0.180 & 0.302 & 0.173 & 0.296 & \underline{0.172} & \underline{0.295} & \underline{0.172} & \underline{0.295} & 0.179 & 0.301 & 0.186 & 0.305 & 0.181 & 0.303 & \underline{0.172} & 0.316 \\
& 336 &  0.317 & 0.408 & 0.321 & 0.411 & \underline{0.300} & \textbf{0.399} & 0.316 & 0.406 & 0.365 & 0.434 & 0.314 & \underline{0.405} & 0.337 & 0.421 & 0.380 & 0.449 & 0.324 & 0.412 & \textbf{0.272} & 0.407 \\
& 720 &  0.800 & 0.670 & 0.835 & 0.688 & \textbf{0.710} & \textbf{0.633} & 0.914 & 0.733 & 0.965 & 0.742 & 0.821 & 0.682 & 0.852 & 0.696 & 0.896 & 0.712 & 0.838 & 0.687 & \underline{0.714} & \underline{0.658} \\
& Avg &  0.340 & \underline{0.391} & 0.355 & 0.401 & \underline{0.319} & \textbf{0.384} & 0.371 & 0.409 & 0.396 & 0.418 & 0.347 & 0.395 & 0.363 & 0.406 & 0.388 & 0.419 & 0.357 & 0.402 & \textbf{0.315} & 0.404 \\
\midrule

\multirow{5}{*}{\rotatebox[origin=c]{90}{Electricity}}
& 96  &  0.148 & 0.239 & \underline{0.136} & 0.233 & 0.145 & \underline{0.232} & \textbf{0.130} & \textbf{0.227} & 0.152 & 0.243 & 0.209 & 0.280 & 0.147 & 0.239 & 0.147 & 0.245 & 0.199 & 0.277 & 0.164 & 0.269 \\
& 192 &  0.162 & 0.251 & 0.164 & 0.258 & \underline{0.160} & \underline{0.246} & \textbf{0.150} & \textbf{0.244} & 0.166 & 0.255 & 0.205 & 0.281 & 0.165 & 0.257 & \underline{0.160} & 0.250 & 0.199 & 0.280 & 0.177 & 0.285 \\
& 336 &  0.180 & 0.270 & 0.183 & 0.279 & 0.174 & \textbf{0.261} & \textbf{0.166} & \underline{0.262} & 0.183 & 0.274 & 0.218 & 0.295 & 0.177 & 0.270 & \underline{0.173} & 0.267 & 0.213 & 0.295 & 0.193 & 0.304 \\
& 720 &  0.217 & 0.303 & 0.213 & 0.303 & \underline{0.203} & \textbf{0.287} & \textbf{0.200} & \underline{0.292} & 0.226 & 0.311 & 0.260 & 0.327 & 0.209 & 0.298 & 0.210 & 0.309 & 0.255 & 0.327 & 0.212 & 0.321 \\
& Avg &  0.176 & \underline{0.265} & 0.174 & 0.268 & \underline{0.170} & \textbf{0.256} & \textbf{0.161} & \textbf{0.256} & 0.181 & 0.270 & 0.223 & 0.295 & 0.174 & 0.266 & 0.172 & 0.267 & 0.216 & 0.294 & 0.186 & 0.294 \\
\midrule

\multicolumn{2}{c|}{1st count}
& \multicolumn{2}{c|}{\textbf{46}}
& \multicolumn{2}{c|}{1}
& \multicolumn{2}{c|}{\underline{14}}
& \multicolumn{2}{c|}{8}
& \multicolumn{2}{c|}{5}
& \multicolumn{2}{c|}{1}
& \multicolumn{2}{c|}{0}
& \multicolumn{2}{c|}{2}
& \multicolumn{2}{c|}{1}
& \multicolumn{2}{c}{1}
\\
\bottomrule
\end{tabular}

} 
\end{table*}

\subsection{Loss Function}

To guarantee accurate forecasting while enforcing structural coherence across the scales, we propose a composite objective function. The total loss $\mathcal{L}_{total}$ combines a primary forecasting loss with a multi-scale regularization term:
\begin{equation}
\mathcal{L}_{total} = \mathcal{L}_{pred} + \alpha \cdot \mathcal{L}_{scale},
\end{equation}
where $\alpha$ is a hyperparameter balancing the two terms.
\paragraph{Primary Forecasting Loss.} To optimize the final prediction accuracy, we employ the MSE loss to minimize the discrepancy between the fused output $\hat{\mathbf{Y}}^{final}$ and the ground truth $\mathbf{Y}^{(0)}$ at the original scale:
\begin{equation}
\mathcal{L}_{pred} = \text{MSE}(\hat{\mathbf{Y}}^{final}, \mathbf{Y}^{(0)}).
\end{equation}

\paragraph{Scale Loss.} However, supervising only the final output is often insufficient to drive deep multi-scale branches to learn effective representations, which can lead to intermediate-scale degeneration or an over-reliance on the fusion layer for error correction. To address this, we introduce a multi-scale regularization term $\mathcal{L}_{scale}$ as a structured constraint. This term comprises two key components: multi-scale supervision, which enforces prediction accuracy at each individual scale to prevent branch degeneration, and cross-scale consistency, which aligns coarse-scale predictions with interval-averaged fine-scale outputs. Together, they exploit the intrinsic multi-scale structure of time series to promote consistent learning across the scales. Formally, the scale loss is defined as:
\begin{equation}
\mathcal{L}_{scale} = \lambda \cdot \mathcal{L}_{superv} + (1-\lambda) \cdot \mathcal{L}_{consist},
\end{equation}
where $\lambda$ is an adaptive weight balancing the two losses:
\begin{equation}
\lambda = \sigma \left( 2 \cdot \left( \frac{\mathcal{L}_{superv}}{\mathcal{L}_{superv} + \mathcal{L}_{consist} + \epsilon} - 0.5 \right) \right),
\end{equation}
where $\epsilon$ is a small constant. This mechanism automatically adjusts the optimization focus, preventing one loss term from dominating the gradient updates and ensuring a stable trade-off between accurate data fitting and structural consistency.
\begin{itemize}
\item \textbf{Multi-scale Supervision.} Given a multi-scale target set $\{\mathbf{Y}^{(0)},\mathbf{Y}^{(1)},\ldots,\mathbf{Y}^{(S)}\}$, the loss imposes constraints on each scale branch to ensure they learn effective representations independently:
\begin{equation}
\mathcal{L}_{superv} = \frac{1}{S+1}\sum\nolimits_{s=0}^{S}\text{MSE}(\hat{\mathbf{Y}}^{(s)}, \mathbf{Y}^{(s)}),
\end{equation}
where $\mathbf{Y}^{(s)}$ is the downsampled ground truth at scale $s$ with length $L_s$. 
\item \textbf{Cross-scale Consistency.} Simultaneously, Cross-scale Consistency loss aligns predictions across scales by enforcing that a fine-scale prediction $\hat{\mathbf{Y}}^{(i)}$, after downsampling, matches its coarse-scale counterpart $\hat{\mathbf{Y}}^{(j)}$ ($i < j$):
\begin{equation}
    \mathcal{L}_{consist} = \frac{2}{S(S+1)}\sum_{i<j} \text{MSE}(\hat{\mathbf{Y}}^{(i)\prime}, \hat{\mathbf{Y}}^{(j)}),
    \label{eq:consist_loss}
\end{equation}
where $\hat{\mathbf{Y}}^{(i)\prime}$ is obtained by downsampling $\hat{\mathbf{Y}}^{(i)}$ to match the scale of $\hat{\mathbf{Y}}^{(j)}$.
\end{itemize}

\section{Experiments}

\subsection{Experimental Setup}


\paragraph{\textbf{Datasets.}}
We evaluate MWMixer on seven widely used long-term forecasting benchmarks, including four ETT datasets (ETTh1, ETTh2, ETTm1, and ETTm2), Weather, Electricity, and Exchange. The ETT datasets record seven electricity transformer variables from July 2016 to July 2018, with hourly sampling for ETTh1/ETTh2 and 15-minute sampling for ETTm1/ETTm2. Weather contains 21 meteorological variables collected every 10 minutes in 2020; Electricity records hourly electricity consumption from 321 clients; and Exchange contains daily exchange rates from eight countries. Following prior work~\cite{wang2023timemixer}, we split the four ETT datasets into training/validation/test sets with a 6:2:2 ratio, and use a 7:1:2 split for the remaining datasets. Dataset statistics are summarized in Table~\ref{tab:dataset_description}.

\begin{table}[H]
  \centering
  \caption{Dataset statistics.}
  \label{tab:dataset_description}
  \begin{tabular}{cccc}
    \toprule
    Dataset & Timesteps & Sample Frequency & Dimension \\
    \midrule
    ETTh1 & 17,420 & 1 hour & 7 \\
    ETTh2 & 17,420 & 1 hour & 7 \\
    ETTm1 & 69,680 & 15 min & 7 \\
    ETTm2 & 69,680 & 15 min & 7 \\
    Weather & 52,696 & 10 min & 21 \\
    Electricity & 26,304 & 1 hour & 321 \\
    Exchange & 7,588 & 1 day & 8 \\
    \bottomrule
  \end{tabular}
\end{table}


\begin{table*}[t!]
  \centering
  \caption{Ablation study of BFBM, the Scale Loss, and their internal branches. The best results are highlighted in \textbf{bold}.}
  \label{tab:ablation_study}
    \begin{tabular}{@{}cc|*{6}{cc|}cc@{}}
    \toprule
    \multicolumn{2}{c|}{Method}
    & \multicolumn{2}{c|}{Ours}
    & \multicolumn{2}{c|}{w/o BFBM}
    & \multicolumn{2}{c|}{w/o DetailMix}
    & \multicolumn{2}{c|}{w/o ApproxMix}
    & \multicolumn{2}{c|}{w/o Scale Loss}
    & \multicolumn{2}{c|}{w/o Supervision}
    & \multicolumn{2}{c}{w/o Consistency} \\
    \cmidrule{1-2}\cmidrule{3-16}
    \multicolumn{2}{c|}{Metric}
    & MSE & MAE & MSE & MAE & MSE & MAE & MSE & MAE
    & MSE & MAE & MSE & MAE & MSE & MAE \\
    \midrule

    \multirow{5}{*}{\rotatebox[origin=c]{90}{ETTh1}}
      & 96  & 0.368 & 0.389 & 0.372 & \textbf{0.387} & 0.369 & 0.389 & 0.372 & \textbf{0.387} & \textbf{0.367} & 0.388 & 0.368 & 0.388 & 0.368 & 0.389 \\
      & 192 & 0.416 & 0.417 & 0.424 & 0.417 & \textbf{0.415} & \textbf{0.416} & 0.428 & 0.420 & \textbf{0.415} & 0.419 & 0.424 & 0.417 & 0.416 & 0.417 \\
      & 336 & \textbf{0.450} & \textbf{0.436} & 0.479 & 0.445 & 0.474 & 0.444 & 0.482 & 0.444 & 0.466 & 0.441 & 0.479 & 0.442 & \textbf{0.450} & \textbf{0.436} \\
      & 720 & \textbf{0.440} & \textbf{0.449} & 0.495 & 0.471 & 0.442 & \textbf{0.449} & 0.491 & 0.468 & 0.460 & 0.460 & 0.479 & 0.468 & 0.460 & 0.460 \\
      & Avg & \textbf{0.418} & \textbf{0.422} & 0.442 & 0.430 & 0.425 & 0.424 & 0.443 & 0.429 & 0.427 & 0.427 & 0.437 & 0.428 & 0.423 & 0.425 \\
    \midrule

    \multirow{5}{*}{\rotatebox[origin=c]{90}{ETTm1}}
      & 96  & \textbf{0.310} & \textbf{0.351} & 0.318 & 0.357 & 0.320 & 0.358 & 0.311 & 0.353 & 0.312 & 0.352 & 0.311 & 0.352 & \textbf{0.310} & \textbf{0.351} \\
      & 192 & \textbf{0.353} & \textbf{0.376} & 0.361 & 0.379 & 0.360 & 0.379 & 0.355 & 0.378 & \textbf{0.353} & \textbf{0.376} & 0.357 & 0.377 & 0.354 & 0.377 \\
      & 336 & 0.387 & \textbf{0.400} & 0.393 & 0.401 & 0.391 & 0.402 & 0.387 & \textbf{0.400} & \textbf{0.385} & \textbf{0.400} & 0.391 & \textbf{0.400} & 0.388 & \textbf{0.400} \\
      & 720 & \textbf{0.443} & \textbf{0.435} & 0.456 & 0.437 & \textbf{0.443} & \textbf{0.435} & 0.456 & 0.437 & \textbf{0.443} & \textbf{0.435} & 0.452 & 0.436 & \textbf{0.443} & \textbf{0.435} \\
      & Avg & \textbf{0.373} & \textbf{0.390} & 0.382 & 0.393 & 0.378 & 0.393 & 0.377 & 0.392 & \textbf{0.373} & \textbf{0.390} & 0.377 & 0.391 & \textbf{0.373} & \textbf{0.390} \\
    \midrule

    \multirow{5}{*}{\rotatebox[origin=c]{90}{Exchange}}
      & 96  & \textbf{0.081} & \textbf{0.198} & \textbf{0.081} & \textbf{0.198} & 0.084 & 0.201 & 0.083 & 0.200 & \textbf{0.081} & \textbf{0.198} & \textbf{0.081} & \textbf{0.198} & \textbf{0.081} & \textbf{0.198} \\
      & 192 & \textbf{0.165} & \textbf{0.289} & 0.179 & 0.300 & 0.178 & 0.299 & 0.179 & 0.299 & 0.194 & 0.311 & 0.194 & 0.311 & 0.177 & 0.298 \\
      & 336 & \textbf{0.317} & 0.408 & 0.342 & 0.422 & 0.343 & 0.423 & 0.339 & 0.419 & 0.318 & 0.408 & 0.357 & 0.432 & \textbf{0.317} & \textbf{0.407} \\
      & 720 & 0.800 & \textbf{0.670} & 0.923 & 0.720 & 0.907 & 0.718 & 0.889 & 0.709 & \textbf{0.799} & \textbf{0.670} & 0.934 & 0.715 & 0.801 & 0.671 \\
      & Avg & \textbf{0.340} & \textbf{0.391} & 0.381 & 0.410 & 0.378 & 0.410 & 0.372 & 0.406 & 0.348 & 0.396 & 0.391 & 0.414 & 0.344 & 0.393 \\
    \bottomrule
    \end{tabular}%
\end{table*}

\paragraph{\textbf{Baselines.}} We evaluate MWMixer against nine state-of-the-art baselines, categorized into three groups: (1) Multi-scale models, including AMD~\cite{hu2025amd}, TimeMixer~\cite{wang2023timemixer}, and MICN~\cite{micn}; (2) MLP-based models, such as FilterNet~\cite{yi2024filternet}, FITS~\cite{xu2024fits}, and SparseTSF~\cite{lin2024sparsetsf}; and (3) Transformer-based models, including CASA~\cite{casa}, DUET~\cite{qiu2025duet}, and iTransformer~\cite{liu2024itransformer}.


\paragraph{\textbf{Implementation Details.}}
We report forecasting performance using MSE and MAE. All experiments are implemented in PyTorch and optimized with Adam using an initial learning rate of $10^{-4}$. The batch size is set to 32 for Electricity and Exchange, and 128 for the other datasets. The model dimension $d$ is selected from $\{64, 128, 256, 512, 1024\}$; the number of downsampling layers $S$, MODWT decomposition levels $M$, and BFBM stack layers $K$ are selected from $\{1, 2, 3\}$. The MODWT wavelet type is selected from $\{\mathrm{haar}, \mathrm{db2}, \mathrm{db3}, \mathrm{db5}, \mathrm{bior3.1}, \mathrm{coif4}, \mathrm{coif5}\}$, the dropout rate is tuned within $[0.0, 0.5]$, and the loss coefficient $\alpha$ is selected from $\{0.001, 0.01, 0.1, 0.3, 0.5, 1.0, 10.0\}$. Training is conducted for at most 30 epochs with early stopping patience of 10 epochs. For fair comparison, all baseline models are configured according to their official implementations when available. Each experiment is repeated three times, and the average performance is reported. 
For reproducibility, the source code of MWMixer is available at:
\url{https://github.com/xsy413/MWMixer}.

\subsection{Main Forecasting Results}
Table \ref{tab:all_results} presents long-term forecasting results under four prediction horizons. MWMixer achieves the best result on 46 out of 70 metrics and remains competitive on the remaining datasets, demonstrating strong overall performance across diverse temporal patterns. Compared with representative multi-scale models such as AMD and TimeMixer, MWMixer benefits from preserving frequency-localized representations during multi-scale modeling. On Electricity, AMD obtains stronger results on several metrics, likely because this dataset contains hundreds of client-level variables and AMD explicitly models inter-variable dependencies. Nevertheless, MWMixer remains best or highly competitive on most benchmarks, supporting the effectiveness of combining MODWT-based decomposition with cross-scale frequency-band mixing.

\subsection{Ablation Studies}
\paragraph{Component Ablation.}
To rigorously evaluate the contribution of each component within our proposed framework, we conducted a comprehensive ablation study. We fixed the look-back window at $L = 96$ and reported average metrics across four horizons ($H \in \{96, 192, 336, 720\}$). We categorized the ablation variants into three distinct groups to analyze the model from the perspectives of multi-scale mixing, dynamic fusion, and scale loss, respectively. 

For BFBM, Table \ref{tab:ablation_study} shows that removing the module leads to clear performance degradation, particularly on Exchange, confirming the importance of cross-scale interaction. Regarding internal branches, removing the approximation path w/o ApproxMix increases errors, indicating that coarse-scale trends are useful for guiding fine-scale learning. Similarly, removing the detail path w/o DetailMix degrades performance, validating that explicit detail injection helps recover information lost during downsampling. These results show that the two directional paths provide complementary support for MWMixer.

We further compare BFBM with common cross-scale mixing strategies in Table~\ref{tab:alternative_strategy_avg}, including Direct Add, Residual Add, Concat+Linear, and Concat+MLP. Direct addition tends to mix features with different frequency properties too rigidly, which can disturb the representation when coarse components and fine-scale component have different energy levels. Although concatenation-based variants introduce learnable transformations after feature concatenation, they still merge multi-scale coefficients without explicitly distinguishing approximation and detail components. This may dilute information or bias the projected representation toward a dominant component. In contrast, BFBM performs structured bidirectional interactions between approximation and detail bands, leading to stronger overall performance.

For Scale Loss, Table \ref{tab:ablation_study} shows that the supervision and consistency terms provide dataset-dependent but generally beneficial regularization. Removing the supervision term weakens the model more clearly on some datasets, such as Exchange, while the consistency term provides additional structural alignment across scales. Although the gains are not uniform across all datasets and horizons, these results suggest that Scale Loss helps multi-scale branches benefit from both scale-specific guidance and cross-scale coherence.

\begin{figure}[t]
  \centering
  \includegraphics[width=\columnwidth]{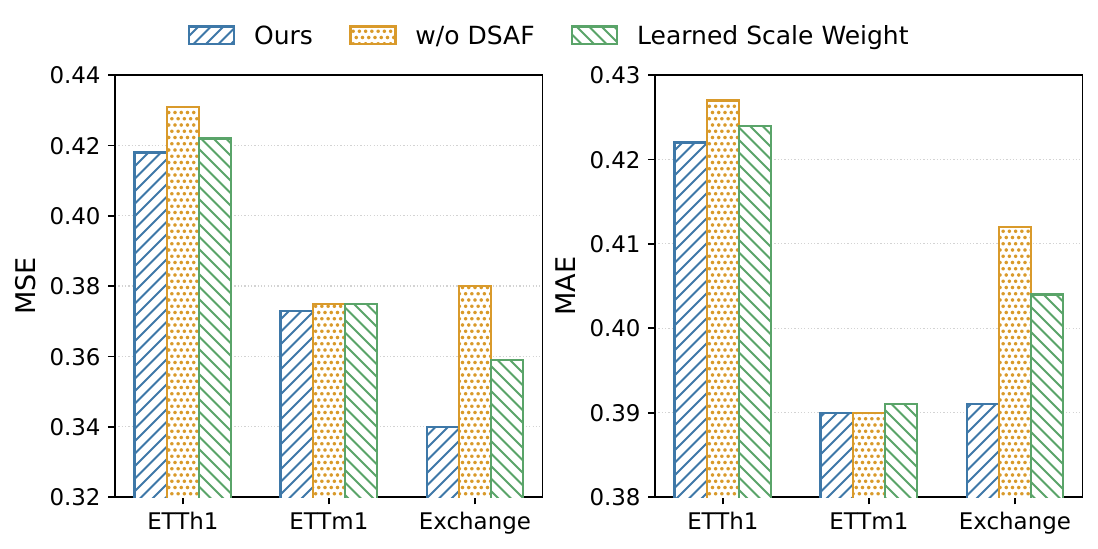}
  \caption{Ablation study on DSAF and learned scale weighting using average results.}
  \Description{A comparison of average forecasting errors for the complete model and variants using different scale-fusion strategies. The plot contrasts Dynamic Scale-Adaptive Fusion with alternatives that remove adaptive fusion or replace learned time-varying scale weights. Lower MSE and MAE values indicate better forecasting performance.}
  \label{fig:ablation_dsaf}
\end{figure}

For DSAF, Figure \ref{fig:ablation_dsaf} and Table \ref{tab:dsaf_fusion_ablation} show that replacing adaptive fusion with static average summation (w/o DSAF) or learned scale weights degrades performance. The gap is especially clear on Exchange, where scale reliability varies more strongly across horizons. This supports the need for time-varying scale aggregation when the reliability of different scales changes across the forecasting horizon.
Detailed results are presented in Appendix B, Table \ref{tab:dsaf_fusion_ablation}.

\paragraph{Decomposition Strategies.}
We investigate decomposition strategies in Table \ref{tab:alternative_strategy_full}. Standard DWT produces consistently higher errors than MODWT across ETTh1, ETTm1, and Exchange, with particularly large degradation on ETTh1 and Exchange. This indicates that the sequence-length reduction in DWT can harm forecasting performance by lowering the temporal resolution of decomposed coefficients. Fourier decomposition is more competitive than DWT, but its averaged results are still generally worse than MODWT. Since Fourier bases provide a global frequency representation, they may be less flexible for non-stationary series with local fluctuations and time-varying patterns. By preserving full sequence length and producing localized time-frequency coefficients, MODWT offers a more suitable decomposition strategy for the multi-scale forecasting framework.

\begin{table*}[!t]
\centering
\caption{Average results of cross-scale mixing and upsampling/decomposition strategies over four prediction horizons $H \in \{96,192,336,720\}$. Full results are provided in Table~\ref{tab:alternative_strategy_full} in Appendix B. The best results are highlighted in \textbf{bold}.}
\label{tab:alternative_strategy_avg}
\resizebox{\textwidth}{!}{%
\begin{tabular}{@{}c|*{9}{cc}@{}}
\toprule
\multicolumn{1}{c|}{Method}
& \multicolumn{2}{c}{Ours}
& \multicolumn{2}{c}{Direct Add}
& \multicolumn{2}{c}{Residual Add}
& \multicolumn{2}{c}{Concat + Linear}
& \multicolumn{2}{c}{Concat + MLP}
& \multicolumn{2}{c}{Nearest}
& \multicolumn{2}{c}{MLP Ups.}
& \multicolumn{2}{c}{DWT}
& \multicolumn{2}{c}{Fourier} \\
\cmidrule{1-19}
\multicolumn{1}{c|}{Metric}
& MSE & MAE
& MSE & MAE
& MSE & MAE
& MSE & MAE
& MSE & MAE
& MSE & MAE
& MSE & MAE
& MSE & MAE
& MSE & MAE \\
\midrule
ETTh1
& 0.418 & \textbf{0.422}
& 0.441 & 0.431
& 0.443 & 0.430
& 0.438 & 0.428
& 0.432 & 0.430
& \textbf{0.417} & \textbf{0.422}
& 0.432 & 0.430
& 0.596 & 0.518
& 0.440 & 0.433 \\
ETTm1
& \textbf{0.373} & \textbf{0.390}
& 0.377 & 0.392
& 0.378 & 0.391
& 0.380 & 0.392
& 0.379 & 0.396
& \textbf{0.373} & \textbf{0.390}
& 0.374 & 0.391
& 0.430 & 0.420
& 0.377 & 0.392 \\
Exchange
& \textbf{0.340} & \textbf{0.391}
& 0.383 & 0.411
& 0.373 & 0.407
& 0.364 & 0.402
& 0.380 & 0.412
& 0.348 & 0.397
& 0.373 & 0.406
& 0.422 & 0.444
& 0.357 & 0.398 \\
\bottomrule
\end{tabular}%
}
\end{table*}

\paragraph{Upsampling Strategies.}
Table \ref{tab:alternative_strategy_avg} compares different upsampling strategies for aligning coarse-scale predictions. Linear and nearest interpolation achieve similar results on the ETT datasets, suggesting that simple non-parametric alignment is sufficient in these relatively regular settings. On Exchange, however, nearest interpolation degrades more clearly at longer horizons, while MLP-based upsampling also introduces larger errors in most cases. We therefore adopt linear interpolation as a stable default, as it provides balanced average performance without adding extra learnable parameters.

\subsection{Sensitivity Analysis}
\begin{figure}[t]
  \centering
  \includegraphics[width=\columnwidth]{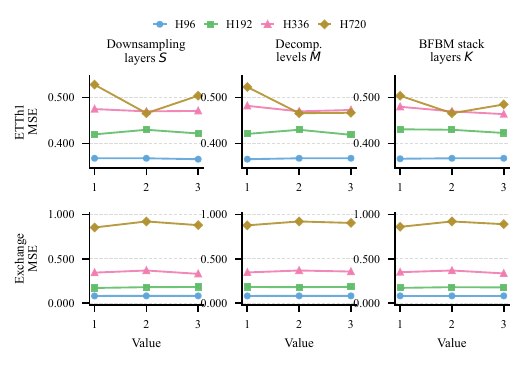}
  \caption{Hyperparameter sensitivity analysis on ETTh1 and Exchange. Each subplot reports MSE under different values of one hyperparameter, while the other two hyperparameters are fixed.}
  \Description{A set of sensitivity plots for ETTh1 and Exchange. In each subplot, the horizontal axis varies one model hyperparameter while the remaining hyperparameters are fixed, and the vertical axis reports mean squared error. The plots show how forecasting performance changes across the tested settings and identify ranges in which the model remains stable.}
  \label{fig:skm_sensitivity}
\end{figure}

We analyze the sensitivity of MWMixer to the number of downsampling layers $S$, MODWT decomposition levels $M$, and stacked BFBM layers $K$. We vary one hyperparameter at a time and evaluate $S$, $M$, and $K$ in $\{1,2,3\}$. Figure~\ref{fig:skm_sensitivity} reports the MSE results on ETTh1 and Exchange under four prediction horizons.
Overall, MWMixer shows stable performance across different hyperparameter choices. On ETTh1, moderate settings generally reduce long-horizon errors, indicating that suitable scale construction and frequency decomposition benefit long-term temporal modeling. On Exchange, longer horizons naturally yield larger absolute MSE values, while the curves remain smooth across different hyperparameter values. This shows that MWMixer does not rely on a narrow hyperparameter configuration to maintain reasonable performance. These results indicate that moderate settings provide a robust default choice, while validation-based selection can further adapt the model to datasets with different temporal characteristics.

\begin{figure}[!tbp]
  \centering
  \includegraphics[width=\columnwidth]{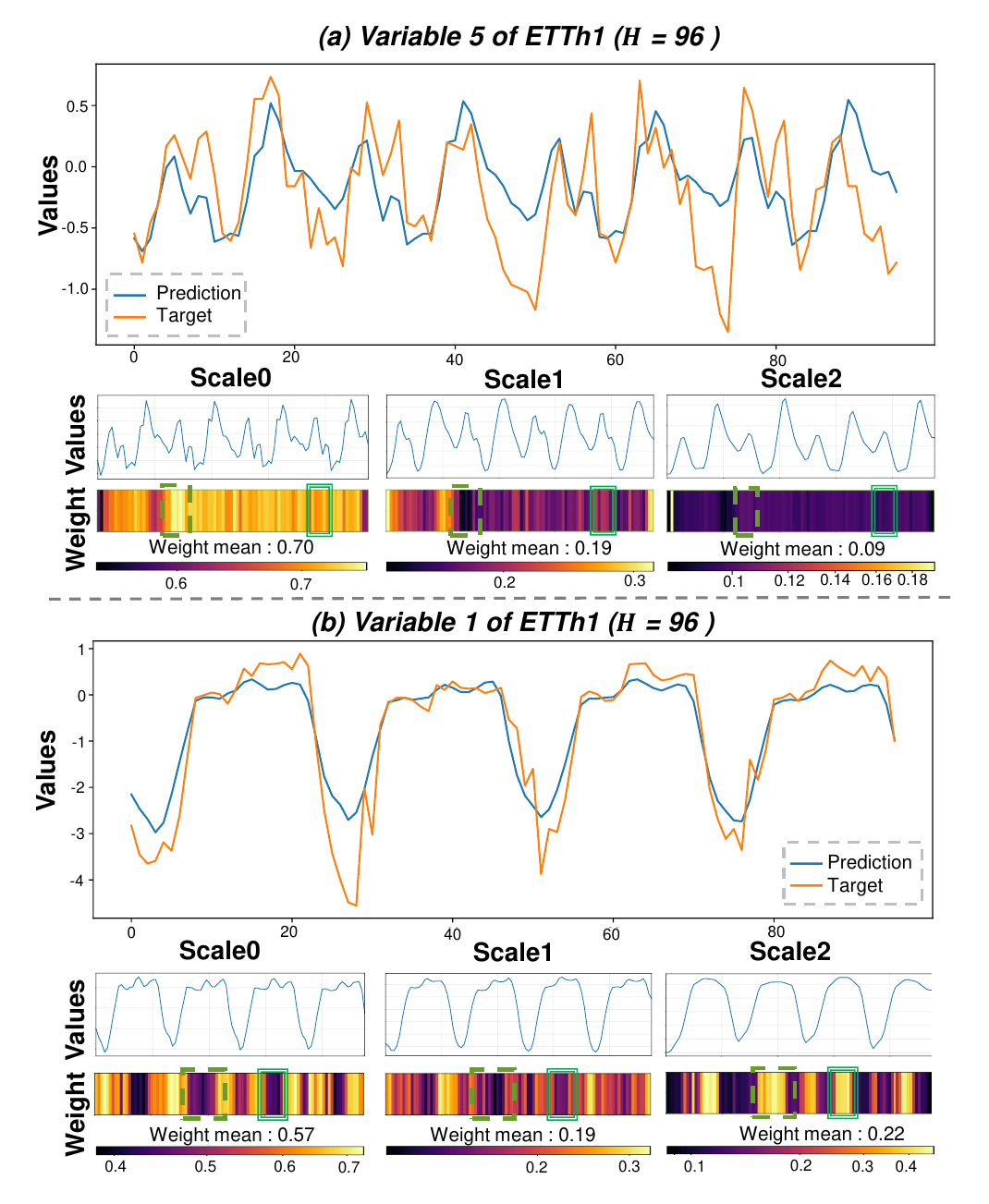}
  \caption{Visualization of adaptive weight allocation by the DSAF module on the ETTh1 dataset for (a) Variable 5 and (b) Variable 1. Top panels display the comparison between predictions and ground truth, while bottom panels show the multi-scale prediction sequences and the time-varying fusion weights learned by the model.}
  \Description{Two groups of time-series plots for variables 5 and 1 of ETTh1. The upper plot in each group overlays the final prediction and ground truth over the forecast horizon. The lower plots display predictions from individual temporal scales together with the fusion weights assigned to those scales at each time step, illustrating that DSAF changes its scale allocation over time and across variables.}
  \label{fig:dsaf_vis}
\end{figure}

\begin{figure*}[t]
  \centering
  
  \begin{minipage}{0.48\textwidth}
    \centering
    \includegraphics[width=\linewidth]{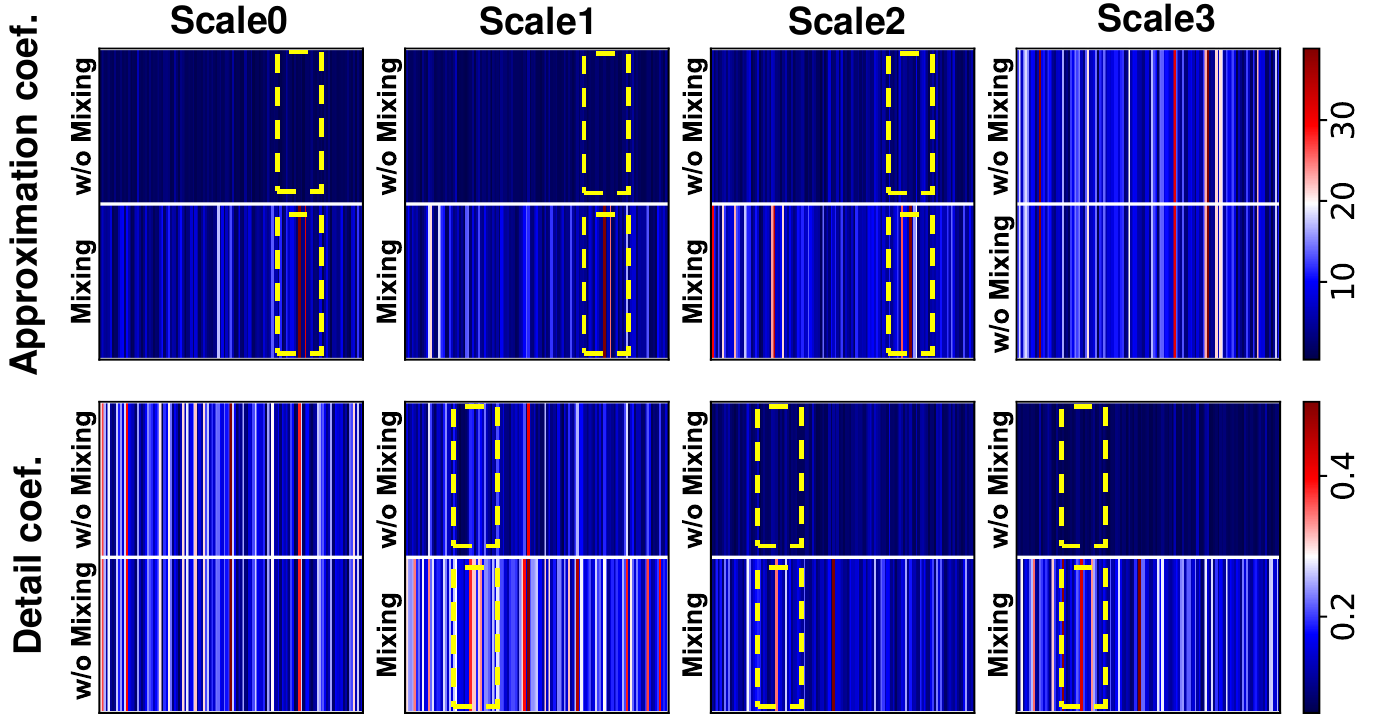}
    \caption{Visualization of energy variations in MODWT coefficients before and after BFBM. The heatmaps illustrate the energy magnitude for Approximation Coefficient and Detail Coefficient.}
    \Description{Heatmaps comparing the energy of MODWT approximation and detail coefficients before and after Bidirectional Frequency-Bands Mixing. Rows and columns organize coefficient bands and temporal positions, and color intensity represents energy magnitude. The post-mixing maps exhibit redistributed and recovered energy across frequency bands and scales.}
    \label{fig:mixvis}
  \end{minipage}
  \hfill
  \begin{minipage}{0.48\textwidth}
    \centering
    \includegraphics[width=\linewidth]{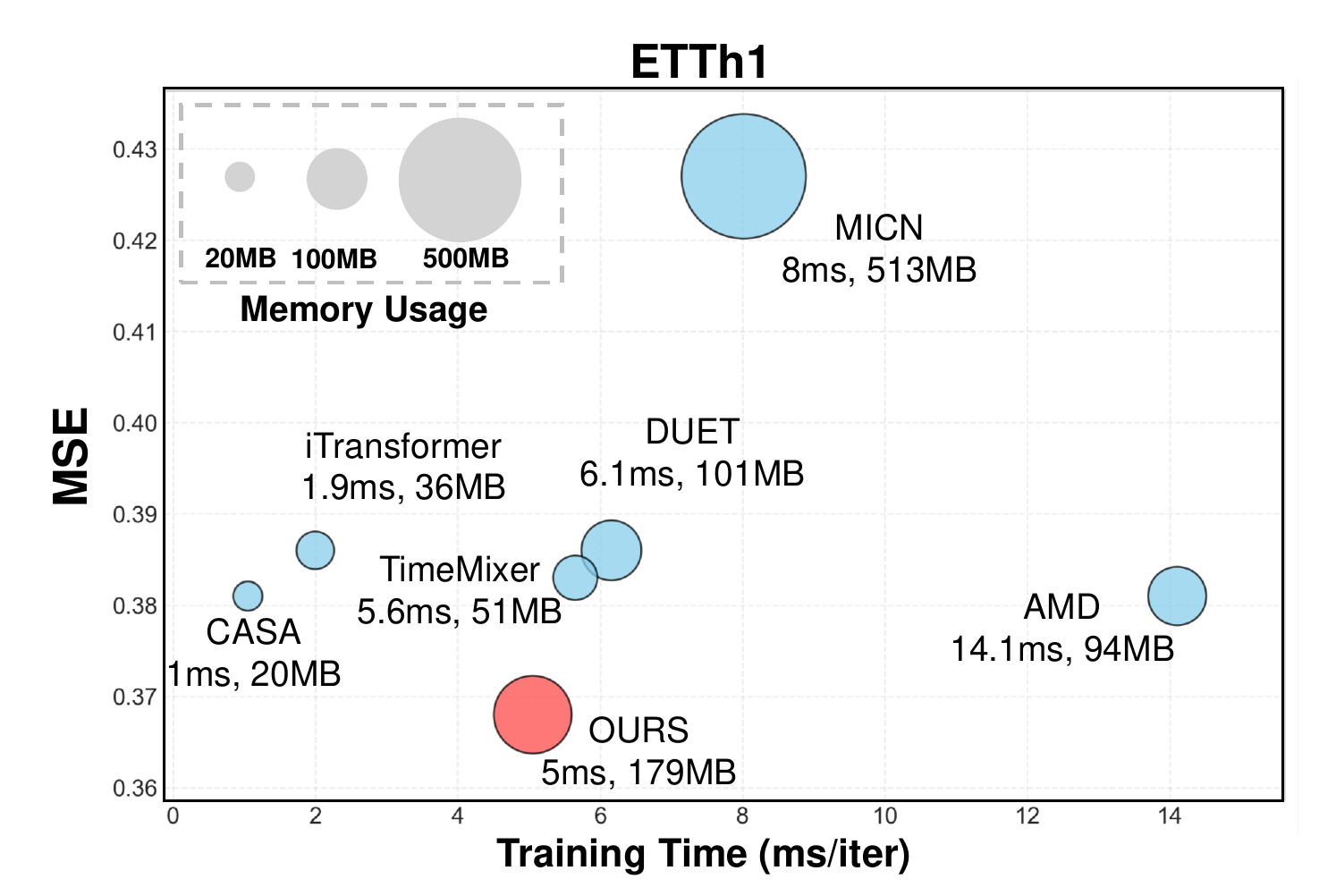}
    \caption{Model efficiency comparison on the ETTh1 dataset. All models are trained with a fixed batch size of 16, and both input and prediction lengths are set to 96.}
    \Description{An efficiency comparison of forecasting models on ETTh1 under identical batch size, input length, and prediction length. The plot compares computational or resource cost with forecasting performance, showing the relative efficiency and accuracy of MWMixer and the baseline models.}
    \label{fig:efficiency}
  \end{minipage}
 
  \par

  \includegraphics[width=0.95\textwidth]{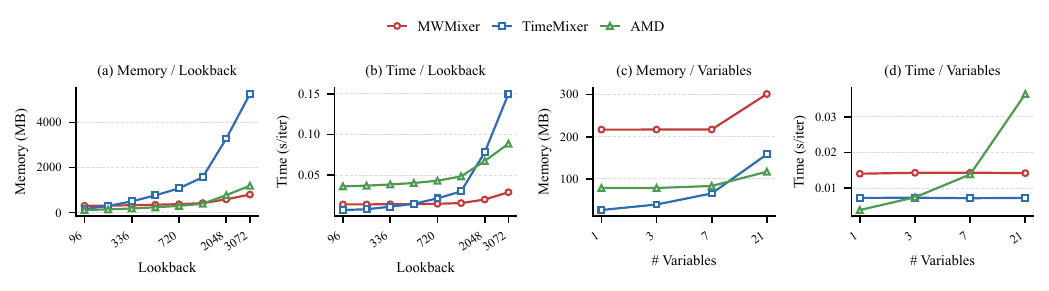}
  \caption{Scalability analysis on Weather. The length analysis fixes the number of variables to 21, while the variable analysis fixes the lookback window to 96.}
  \Description{Scalability plots on the Weather dataset. One analysis varies the sequence length while holding the number of variables at 21, and the other varies the number of variables while holding the lookback window at 96. The plotted resource or runtime measurements compare how MWMixer and baseline models scale as each input dimension increases.}
  \label{fig:eff_scalability}
  
\end{figure*}

\subsection{Visualization and Efficiency Analysis}

\paragraph{Visualization of DSAF}
Figure \ref{fig:dsaf_vis} visualizes the adaptive weight allocation of DSAF. For Variable 5, characterized by high volatility, the model predominantly relies on the fine-grained Scale to capture rapid fluctuations, suppressing the coarse Scale. In contrast, for the strongly periodic Variable 1, the model shifts focus to capture stable trends, significantly increasing the contribution of the coarse Scale 2. This confirms that DSAF dynamically prioritizes fine-grained scales for high-frequency details while leveraging coarse-grained scales for periodicity and trends.

\paragraph{Visualization of BFBM}
Figure \ref{fig:mixvis} visualizes the energy distributions of MODWT approximation and detail coefficients before and after BFBM. In the approximation coefficients, the mixed representations show stronger responses at several highlighted temporal regions on Scales 0--2, indicating that cross-scale interaction enriches trend-related components rather than only preserving the original coarse representation. In the detail coefficients, BFBM also increases localized responses on Scales 1--3, especially in the highlighted regions, showing that fine-grained fluctuations can be propagated to coarser scales after mixing. These observations provide coefficient-level evidence that BFBM adjusts
information flow across both approximation and detail bands, strengthening
trend representations while retaining, rather than uniformly smoothing,
local temporal variations.

\paragraph{Model Efficiency.}
We compared MWMixer with representative multi-scale models. As shown in Figure \ref{fig:efficiency}, MWMixer achieves a favorable accuracy-efficiency trade-off. It is faster than AMD by bypassing computation-heavy adaptive synthesis relying on parallel predictors, and it is lighter than MICN by avoiding heavy multi-scale isometric convolutions. It also matches TimeMixer's speed while delivering stronger forecasting performance.

We further analyze scalability with respect to lookback length and variable dimensionality in Figure~\ref{fig:eff_scalability}. When the lookback window increases, MWMixer shows a much slower growth in memory usage than TimeMixer and remains close to AMD, while its training time increases smoothly and stays lower than both baselines at long lookback lengths. When the number of variables increases, MWMixer keeps nearly constant training time and moderate memory growth. Although TimeMixer is faster along the variable dimension, its memory usage grows more noticeably, whereas AMD exhibits a sharp increase in training time at the largest variable setting. These results show that MWMixer maintains favorable scalability under both longer temporal contexts and higher-dimensional inputs.

\begin{table*}[t]
\centering
\caption{Full results of cross-scale mixing, upsampling, and decomposition strategies. MSE/MAE are reported under four prediction horizons $H \in \{96,192,336,720\}$. The best results are highlighted in \textbf{bold}.}
\label{tab:alternative_strategy_full}
\resizebox{\textwidth}{!}{%
\begin{tabular}{@{}cc|*{9}{cc}@{}}
\toprule
\multicolumn{2}{c|}{Method}
& \multicolumn{2}{c}{Ours} 
& \multicolumn{2}{c}{Direct Add} 
& \multicolumn{2}{c}{Residual Add} 
& \multicolumn{2}{c}{Concat + Linear} 
& \multicolumn{2}{c}{Concat + MLP} 
& \multicolumn{2}{c}{Nearest} 
& \multicolumn{2}{c}{MLP Ups.} 
& \multicolumn{2}{c}{DWT} 
& \multicolumn{2}{c}{Fourier} \\
\cmidrule{1-20}
\multicolumn{2}{c|}{Metric}
& MSE & MAE
& MSE & MAE
& MSE & MAE
& MSE & MAE
& MSE & MAE
& MSE & MAE
& MSE & MAE
& MSE & MAE
& MSE & MAE \\
\midrule
\multirow{5}{*}{\rotatebox[origin=c]{90}{ETTh1}}
& 96
& 0.368 & 0.389
& 0.376 & 0.392
& 0.375 & 0.390
& 0.371 & \textbf{0.388}
& 0.378 & 0.397
& \textbf{0.367} & 0.389
& 0.368 & 0.389
& 0.569 & 0.490
& 0.370 & 0.391 \\
& 192
& 0.416 & \textbf{0.417}
& 0.427 & 0.419
& 0.429 & 0.420
& 0.426 & 0.418
& 0.416 & 0.417
& \textbf{0.415} & 0.417
& 0.416 & 0.417
& 0.582 & 0.502
& 0.426 & 0.424 \\
& 336
& 0.450 & \textbf{0.436}
& 0.481 & 0.445
& 0.484 & 0.445
& 0.473 & 0.441
& 0.481 & 0.448
& \textbf{0.449} & 0.436
& 0.453 & 0.438
& 0.616 & 0.526
& 0.482 & 0.448 \\
& 720
& 0.440 & \textbf{0.449}
& 0.481 & 0.468
& 0.484 & 0.465
& 0.480 & 0.465
& 0.453 & 0.456
& \textbf{0.439} & 0.449
& 0.492 & 0.479
& 0.620 & 0.556
& 0.484 & 0.471 \\
\midrule
\multirow{5}{*}{\rotatebox[origin=c]{90}{ETTm1}}
& 96
& \textbf{0.310} & \textbf{0.351}
& 0.311 & 0.353
& 0.312 & 0.352
& 0.314 & 0.353
& 0.323 & 0.362
& 0.310 & 0.351
& 0.313 & 0.353
& 0.375 & 0.390
& 0.314 & 0.354 \\
& 192
& \textbf{0.353} & \textbf{0.376}
& 0.355 & 0.378
& 0.357 & 0.377
& 0.357 & 0.376
& 0.355 & 0.379
& 0.354 & 0.376
& 0.353 & 0.377
& 0.414 & 0.409
& 0.359 & 0.378 \\
& 336
& 0.387 & 0.400
& 0.387 & 0.400
& 0.388 & \textbf{0.399}
& 0.390 & \textbf{0.399}
& 0.387 & 0.402
& 0.388 & 0.400
& \textbf{0.386} & 0.401
& 0.439 & 0.426
& 0.390 & 0.401 \\
& 720
& \textbf{0.443} & \textbf{0.435}
& 0.454 & 0.438
& 0.454 & 0.437
& 0.459 & 0.439
& 0.452 & 0.442
& 0.443 & 0.435
& 0.445 & 0.435
& 0.493 & 0.457
& 0.445 & 0.435 \\
\midrule
\multirow{5}{*}{\rotatebox[origin=c]{90}{Exchange}}
& 96
& \textbf{0.081} & \textbf{0.198}
& 0.082 & 0.199
& 0.082 & 0.199
& 0.081 & 0.198
& 0.082 & 0.199
& 0.081 & 0.198
& 0.085 & 0.202
& 0.112 & 0.237
& 0.082 & 0.198 \\
& 192
& \textbf{0.165} & \textbf{0.289}
& 0.186 & 0.304
& 0.181 & 0.301
& 0.171 & 0.293
& 0.174 & 0.297
& 0.194 & 0.311
& 0.171 & 0.293
& 0.218 & 0.334
& 0.168 & 0.292 \\
& 336
& \textbf{0.317} & 0.408
& 0.349 & 0.427
& 0.342 & 0.421
& 0.332 & 0.417
& 0.349 & 0.427
& 0.317 & 0.408
& 0.335 & 0.418
& 0.386 & 0.454
& 0.318 & \textbf{0.407} \\
& 720
& \textbf{0.800} & \textbf{0.670}
& 0.914 & 0.715
& 0.888 & 0.707
& 0.871 & 0.700
& 0.916 & 0.723
& 0.801 & 0.671
& 0.902 & 0.712
& 0.974 & 0.753
& 0.861 & 0.697 \\
\bottomrule
\end{tabular}%
}
\end{table*}

\section{Conclusion}
In this paper, we propose MWMixer, a novel framework to address information loss from decomposition and downsampling inherent in existing multi-scale methods. By adopting MODWT, we eliminate decomposition-induced loss, preserving full temporal resolution. Simultaneously, the BFBM module compensates for downsampling-induced loss by explicitly recovering lost details via cross-scale interaction. Furthermore, with the aid of DSAF for adaptive fusion and Scale Loss for structural consistency, MWMixer achieves competitive performance across various benchmarks, offering an effective solution for long-term time series forecasting.

\appendix

\begin{figure*}[t!]
\centering
\begin{subfigure}[b]{0.24\textwidth}
\centering
\includegraphics[width=\textwidth]{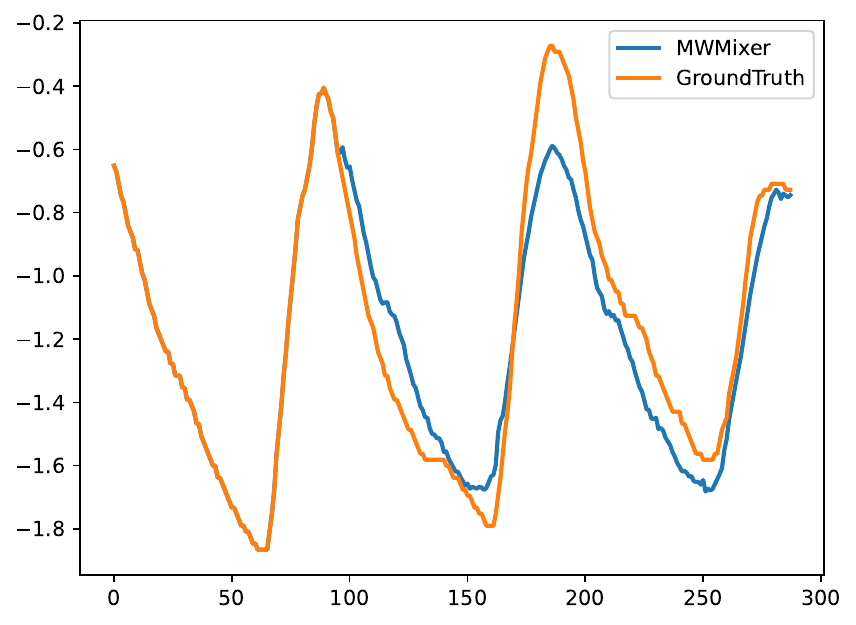}
\caption{MWMixer}
\label{fig:sub_mwmm2}
\end{subfigure}%
\begin{subfigure}[b]{0.24\textwidth}
\centering
\includegraphics[width=\textwidth]{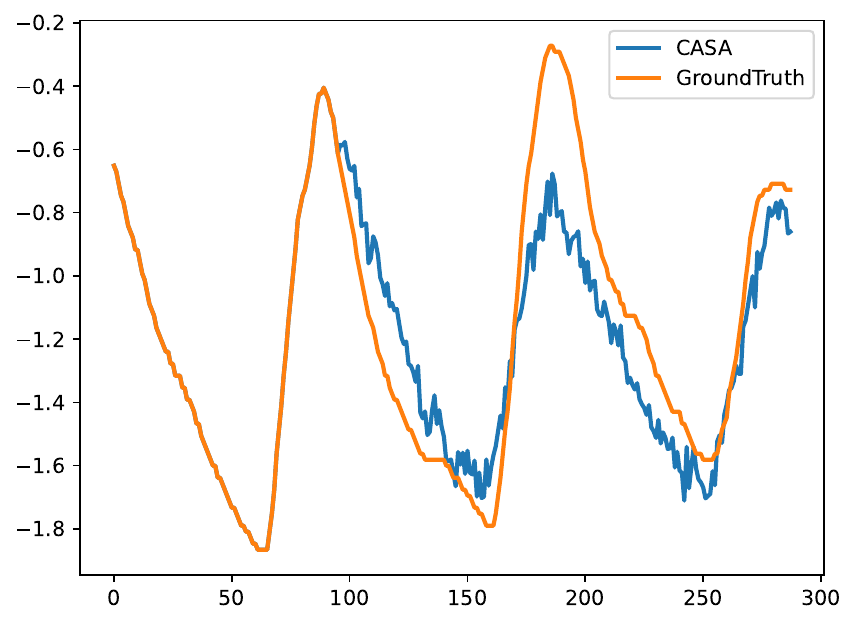}
\caption{CASA}
\label{fig:sub_casam2}
\end{subfigure}%
\begin{subfigure}[b]{0.24\textwidth}
\centering
\includegraphics[width=\textwidth]{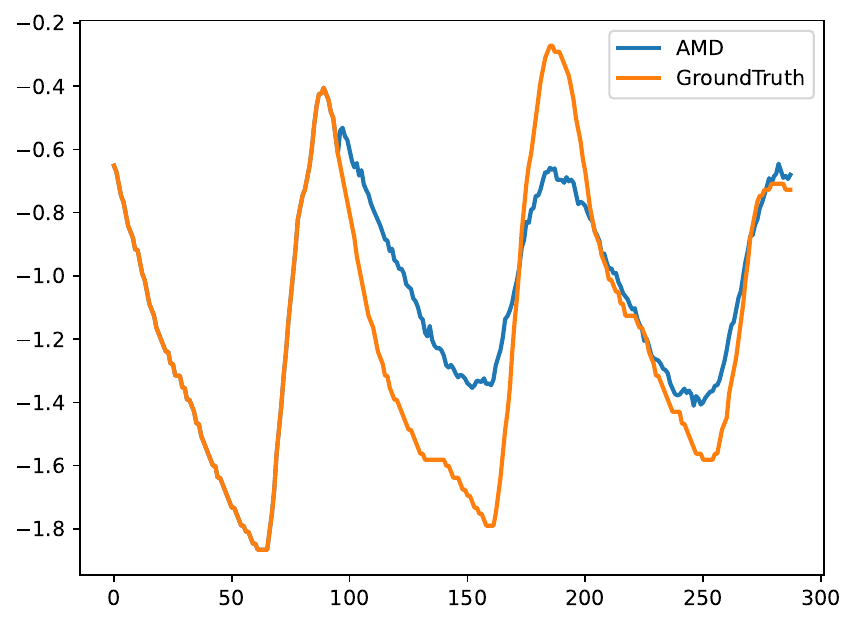}
\caption{AMD}
\label{fig:sub_amdm2}
\end{subfigure}%
\begin{subfigure}[b]{0.24\textwidth}
\centering
\includegraphics[width=\textwidth]{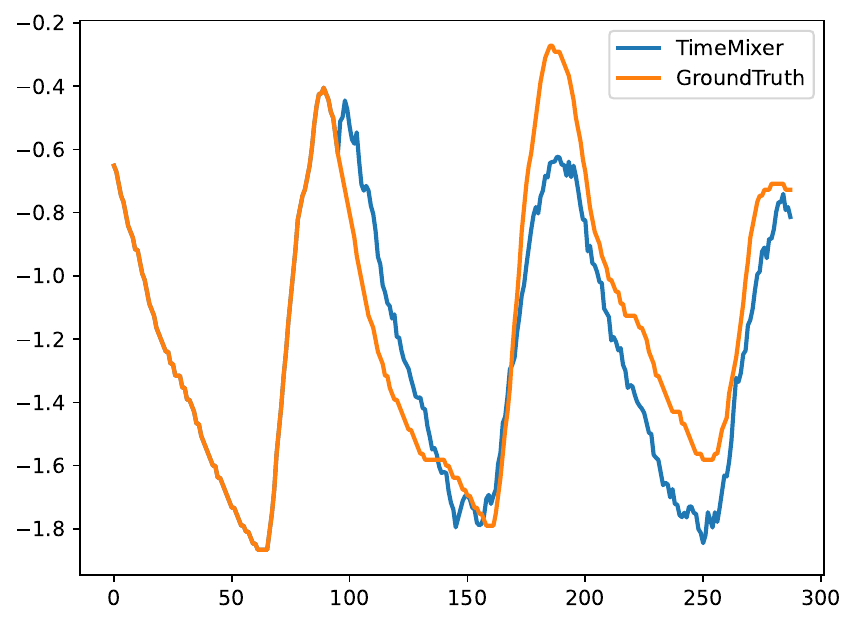}
\caption{TimeMixer}
\label{fig:sub_timemixerm2}
\end{subfigure}
\caption{Prediction Results of Our Method (i.e. MWMixer) and Baseline Models on the ETTm2 Dataset.}
\Description{Four aligned time-series prediction plots on ETTm2 compare MWMixer, CASA, AMD, and TimeMixer over the same forecast interval. Each panel overlays a model prediction with the ground-truth sequence, enabling comparison of how closely each method follows the temporal level, fluctuations, and turning points of the target series.}
\label{fig:visualization_ettm2}
\end{figure*}

\section{Theorem 1 and Its Proof}
\label{app:Theorem 1 and Its Proof}

\begin{theorem}
{(Spectral Separation and Energy Conservation of MODWT).}
Given a time series $\mathbf{X}$, an $M$-level MODWT with normalized
orthogonal filters and periodic boundary extension yields coefficients
$\{\mathbf{D}_1,\ldots,\mathbf{D}_M,\mathbf{A}_M\}$, where
$\mathbf{D}_m$ captures detail components associated with the frequency band
at level $m$, and $\mathbf{A}_M$ represents the residual low-frequency
approximation. Moreover, the decomposition satisfies
\begin{equation}
\|\mathbf{X}\|_F^2
=
\sum_{m=1}^{M}\|\mathbf{D}_m\|_F^2
+
\|\mathbf{A}_M\|_F^2.
\end{equation}
\label{thm:modwt_separation_proof}
\end{theorem}

\noindent\textbf{Proof.}
Let $\mathbf{A}_0=\mathbf{X}$. At level $m$, the MODWT filters yield
\begin{equation}
\mathbf{D}_m=\mathbf{A}_{m-1}*\tilde{h}_m,\qquad
\mathbf{A}_m=\mathbf{A}_{m-1}*\tilde{g}_m,
\end{equation}
where the two branches isolate scale-dependent detail and approximation
components without temporal downsampling. Filter normalization gives
\begin{equation}
|\tilde{H}_m(\omega)|^2+|\tilde{G}_m(\omega)|^2=1,
\end{equation}
and Parseval's identity therefore implies
\begin{equation}
\|\mathbf{A}_{m-1}\|_F^2
=
\|\mathbf{D}_m\|_F^2+\|\mathbf{A}_m\|_F^2.
\end{equation}
Telescoping this identity over $m=1,\ldots,M$ yields the stated energy
decomposition, completing the proof.

\section{Extended Experimental Results}

Table~\ref{tab:alternative_strategy_full} reports full results for
cross-scale mixing, upsampling, and decomposition strategies, while
Table~\ref{tab:dsaf_fusion_ablation} reports DSAF and learned
scale-weighting ablations.

\begin{table}[t]
  \centering
  \caption{Average ablation results for DSAF and learned scale weighting.}
  \label{tab:dsaf_fusion_ablation}
  \begin{tabular}{@{}c|cc|cc|cc@{}}
    \toprule
    Method
    & \multicolumn{2}{c|}{Ours}
    & \multicolumn{2}{c|}{w/o DSAF}
    & \multicolumn{2}{c}{\makecell[c]{Learned Scale\\Weight}} \\
    \midrule
    Metric & MSE & MAE & MSE & MAE & MSE & MAE \\
    \midrule
    ETTh1    & \textbf{0.418} & \textbf{0.422} & 0.431 & 0.427 & 0.422 & 0.424 \\
    ETTm1    & \textbf{0.373} & \textbf{0.390} & 0.375 & \textbf{0.390} & 0.375 & 0.391 \\
    Exchange & \textbf{0.340} & \textbf{0.391} & 0.380 & 0.412 & 0.359 & 0.404 \\
    \bottomrule
  \end{tabular}%
\end{table}

\section{Visualization Results}
Figure~\ref{fig:visualization_ettm2} present a comparison between the ground truth and the predictions generated by MWMixer, CASA, AMD, and TimeMixer. Both the input and prediction lengths are set to 96.

\begin{acks}
This work was supported by the National Key Research and Development
Program of China under Grant 2025YFE0118100, the National Natural
Science Foundation of China under Grant 62376289, and the Natural
Science Foundation for Excellent Young Scholars of Hunan Province
under Grant 2024JJ4069.
\end{acks}

\section*{GenAI Usage Disclosure}
The authors utilized ChatGPT for grammar checking and language polishing of the manuscript. No content was generated by generative AI tools.

\bibliographystyle{ACM-Reference-Format}
\bibliography{MWMixer-CIKM26}

\end{document}